\documentclass[nohyperref]{article}

\usepackage{microtype}
\usepackage{graphicx}
\usepackage{subfigure}
\usepackage{booktabs} 
\usepackage{hyperref}

\usepackage[accepted]{icml2022}

\usepackage{amsmath}
\usepackage{amssymb}
\usepackage{mathtools}
\usepackage{amsthm}
\usepackage{bm}
\usepackage{bbm}
\usepackage{amsfonts}
\usepackage{mathrsfs}
\usepackage{graphicx}
\usepackage{subfigure}
\usepackage{wrapfig}
\usepackage{caption}
\usepackage{multirow}

\usepackage[capitalize,noabbrev]{cleveref}

\theoremstyle{plain}

\theoremstyle{definition}

\theoremstyle{remark}

\newcommand{\ie}{\textit{i}.\textit{e}., }

\newcommand{\etc}{\textit{etc}. }
\newcommand{\tabincell}[2]{\begin{tabular}{@{}#1@{}}#2\end{tabular}}  

\usepackage{xcolor}
\definecolor{cadmiumgreen}{rgb}{0.0, 0.42, 0.24}
\definecolor{cadmiumred}{rgb}{0.89, 0.0, 0.13}

\definecolor{darkergreen}{RGB}{21, 152, 56}

\DeclareMathOperator*{\argmin}{arg\,min}
\usepackage[textsize=tiny]{todonotes}

\icmltitlerunning{LightCLIP: Learning Multi-Level Interaction for Lightweight Vision-Language Models}

\begin{document}

\twocolumn[
\icmltitle{LightCLIP: Learning Multi-Level Interaction for Lightweight Vision-Language Models}

\icmlsetsymbol{equal}{*}

\begin{icmlauthorlist}
\icmlauthor{Ying Nie}{equal,xxx}
\icmlauthor{Wei He}{equal,xxx}
\icmlauthor{Kai Han}{xxx}
\icmlauthor{Yehui Tang}{xxx}
\icmlauthor{Tianyu Guo}{xxx}
\icmlauthor{Fanyi Du}{xxx}
\icmlauthor{Yunhe Wang}{xxx}
\end{icmlauthorlist}
\icmlaffiliation{xxx}{Huawei Noah's Ark Lab}
\icmlcorrespondingauthor{Kai Han}{kai.han@huawei.com}
\icmlcorrespondingauthor{Yunhe Wang}{yunhe.wang@huawei.com}

\vskip 0.3in
]

\printAffiliationsAndNotice{\icmlEqualContribution} 

\begin{abstract}
	Vision-language pre-training like CLIP has shown promising performance on various downstream tasks such as zero-shot image classification and image-text retrieval. Most of the existing CLIP-alike works usually adopt relatively large image encoders like ResNet50 and ViT, while the lightweight counterparts are rarely discussed. In this paper, we propose a multi-level interaction paradigm for training lightweight CLIP models. Firstly, to mitigate the problem that some image-text pairs are not strictly one-to-one correspondence, we improve the conventional global instance-level alignment objective by softening the label of negative samples progressively. Secondly, a relaxed bipartite matching based token-level alignment objective is introduced for finer-grained alignment between image patches and textual words. Moreover, based on the observation that the accuracy of CLIP model does not increase correspondingly as the parameters of text encoder increase, an extra objective of masked language modeling (MLM) is leveraged for maximizing the potential of the shortened text encoder. In practice, an auxiliary fusion module injecting unmasked image embedding into masked text embedding at different network stages is proposed for enhancing the MLM. Extensive experiments show that without introducing additional computational cost during inference, the proposed method achieves a higher performance on multiple downstream tasks. 
\end{abstract}

\section{Introduction} 
In the past few years, Vision-Language Pre-training (VLP) has shown its extraordinary performance on multiple downstream tasks, such as zero-shot classification~\cite{deng2009imagenet,krizhevsky2009learning}, image-text retrieval~\cite{hodosh2013framing, lin2014microsoft} and image captioning~\cite{goyal2017making}. The success can be attributed to several factors: large-scale image-text pairs crawled from the web, well-designed model architectures and suitable pre-training objectives. In general, the existing VLP models can be roughly categorized into two paradigms: single-stream and two-stream. Single-stream models often build the cross-modal fusion module for modeling the fine-grained interactions between image patches and textual words. Although they have achieved promising performance on some tasks, the redundant component design and expensive computational cost restrict their wider applications~\cite{li2019visualbert,li2020oscar,gan2020large,li2020unimo,chen2020uniter}. In contrast, two-stream models decouple the encoders of image and text, and extract the embeddings for image and text separately. The representative two-stream models like CLIP~\cite{radford2021learning} and ALIGN~\cite{jia2021scaling} perform global contrastive learning between image and text on large-scale web-crawled image-text pairs, achieving astounding performance on multiple downstream tasks. 
\begin{figure*}[t]
	\centering
	\includegraphics[width=0.95\linewidth]{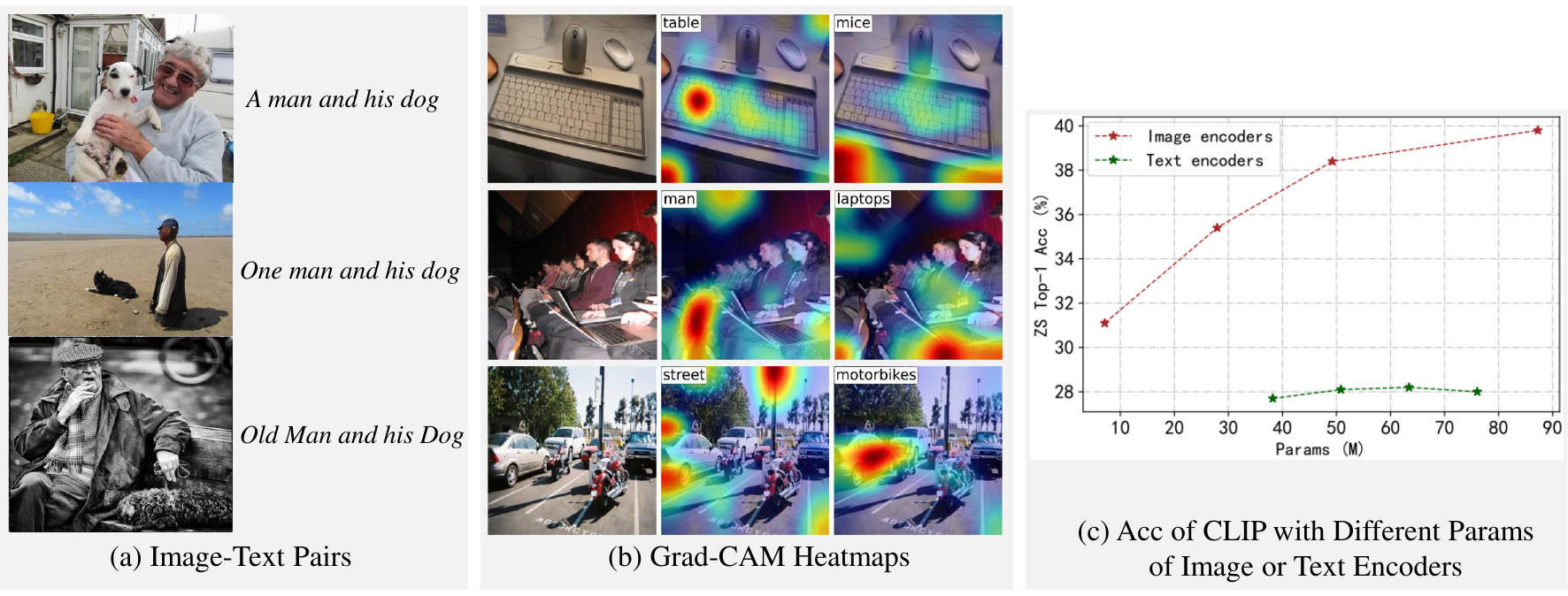}
	\caption{Motivations of our method: (a) Some image-text pairs in YFCC15M-V2 dataset~\cite{li2021supervision} are many-to-many correspondence, indicating the one-hot labels for instance-level alignment are sub-optimal; (b) Adopting only the objective of instance-level alignment usually results in the correspondence failure between image patches and textual words; (c) Under the same text encoder (8-layers Transformers), the ImageNet zero-shot accuracy improves accordingly when Swin-Nano, Swin-Tiny, Swin-Small and Swin-Base~\cite{liu2021swin} are adopted as the image encoder, respectively. However, under the same image encoder (MobileNet-V2), the accuracy does not improve significantly when 4-layer, 8-layer, 12-layer and 16-layer Transformers~\cite{vaswani2017attention,radford2019language} are adopted as the text encoder, respectively.}
	\label{fig0}
	\vspace{-1.0em}
\end{figure*}

Due to its simplicity and astounding accuracy, the two-stream paradigm dominates in the field of multi-modality recently. However, the existing VLP models are difficult to be deployed on edge devices due to its large parameters. For example, most of the existing CLIP-alike works usually adopt ResNet50 and ViT as image encoders~\cite{radford2021learning, gao2022pyramidclip, gao2023softclip, yao2021filip, lu2022cots, li2021supervision, kim2021vilt, li2021align, goel2022cyclip, lee2022uniclip}, while the lightweight counterparts are rarely discussed. The text encoders are mostly composed of 12 transformer blocks with heavy parameters~\cite{gao2022pyramidclip, gao2023softclip, li2021supervision, yao2021filip, yu2022coca}. Therefore, it is an open problem to explore CLIP-alike models equipped with lightweight image and text encoder under affordable computing resources.

Unfortunately, directly copying existing methods to train lightweight CLIP models often yields sub-optimal results. The size-constrained image and text encoder have more difficulty learning the richer representation used to align image and text. Firstly, some web-crawled image-text pairs are not strictly one-to-one correspondence (Figure~\ref{fig0}.(a)), which is more harmful for lightweight CLIP models pre-trained under one-hot labels. Besides, only adopting a global instance-level alignment is effective but not sufficient. For example, the Grad-CAM heatmaps ~\cite{selvaraju2017grad} of  MobileNet-V2~\cite{sandler2018mobilenetv2} image encoder (Figure~\ref{fig0}.(b)) indicate the failure correspondence between image patches and textual words, usually resulting in poorer accuracy. Finally, instead of the mutual disregard between lightweight image and text encoder all the pre-training time, properly fusing the embedding of image and text, \ie mutual help among lightweight image and text encoder may help improve the final performance.

In this paper, we propose LightCLIP which explores an efficient image-text alignment and fusion paradigm for training lightweight CLIP models. To mitigate the problem that some web-crawled image-text pairs are not strictly one-to-one correspondence in pre-training dataset, we first improve the traditional global instance-level alignment objective by softening the labels of negative samples progressively. Considering the limited representation ability of lightweight image and text encoders, we then devise a relaxed bipartite matching based token-level alignment objective for finer-grained alignment between image patches and textual words. Specifically, we view the problem of token-level alignment as a direct set prediction problem, in which the optimal one-to-one correspondence can be computed efficiently with bipartite matching algorithms like Hungarian~\cite{stewart2016end, carion2020end}. Finally, based on the observation that the performance of CLIP model does not increase correspondingly as the parameters of text encoder increase (Figure~\ref{fig0}.(c)), we reduce the number of layers in the text encoder and the objective of masked language modeling (MLM)~\cite{devlin2018bert} is leveraged for helping pre-training. In practice, an auxiliary fusion module is proposed to inject unmasked image embedding into masked text embedding at different network stages for enhancing the MLM. Remarkably, the module of multi-level fusion is only introduced at pre-training time, therefore, the ability to pre-compute image and text embeddings offline can be maintained at inference time. The extensive experiments on multiple benchmarks, and the thorough ablation studies demonstrate the effectiveness of the proposed method. 

\section{Related Work}
Vision-Language Pre-Training (VLP) has achieved promising zero-shot performance on various down-stream tasks by learning an expressive joint representation between two modalities on large-scale image-text pairs. In general, the existing VLP models can be roughly divided into two categories, \ie single-stream and two-stream. The single-stream paradigm models the representation of both image and text using a single deep fusion encoder with cross-modal interaction, such as Visual-BERT~\cite{li2019visualbert}, OSCAR~\cite{li2020oscar}, VILLA~\cite{gan2020large}, UNIMO~\cite{li2020unimo} and UNITER~\cite{chen2020uniter}. Although they have achieved promising performance on some tasks, the cross-modal fusion process on all possible query-candidate pairs inevitably slows down the inference speed. In addition, to extract informative image regions, the off-the-shelf object detector like Faster R-CNN~\cite{girshick2015fast} is typically adopted, which renders the approach less scalable. In contrast, the two-stream paradigm encodes image and text separately with decoupled image and text encoder like ViLBERT~\cite{lu2019vilbert}, CLIP~\cite{radford2021learning} and ALIGN~\cite{jia2021scaling}. Image-text contrastive learning ~\cite{oord2018representation} is usually employed to optimize the image and text encoder simultaneously. 

Due to its simplicity, flexibility, and relatively cheaper computational cost, two-stream VLP models dominate so far. The following CLIP-alike works have been springing up like mushrooms after rain. To mitigate the impact of noisy information in web-crawled image-text pairs, PyramidCLIP~\cite{gao2022pyramidclip}, SoftCLIP~\cite{gao2023softclip} and CLIP-PSD~\cite{andonian2022robust} propose to replace the hard one-hot labels in contrastive learning with softened labels. ALBEF~\cite{li2021align} adopts momentum models to generate pseudo targets as additional supervision. NLIP~\cite{huang2022nlip} proposes a principled noise robust framework including noise-harmonization and noise-completion to stabilize pre-training process. DiHT~\cite{radenovic2023filtering} also improves the performance of VLP from the aspects of dataset noise, model initialization and training objective. Instead of using the common contrastive learning between the global embedding of image and text, the finer token-level alignment is presented in FILIP~\cite{yao2021filip}, COTS~\cite{lu2022cots} and TokenFlow~\cite{zou2022tokenflow} \etc DeCLIP~\cite{li2021supervision} and SLIP~\cite{mu2022slip} further introduce more supervisions to improve data utilization efficiency. CYCLIP~\cite{goel2022cyclip} proposes a novel contrastive learning with two additional cycle consistency constraints for mitigating the problem of inconsistent predictions in image and text domains. However, most of the predecessors' works are keen on relatively-large models, the lightweight models that are more preferred by resources-limited devices are rarely discussed.

\section{Method} 
\begin{figure*}[t]
	\centering
	\includegraphics[width=0.95\linewidth]{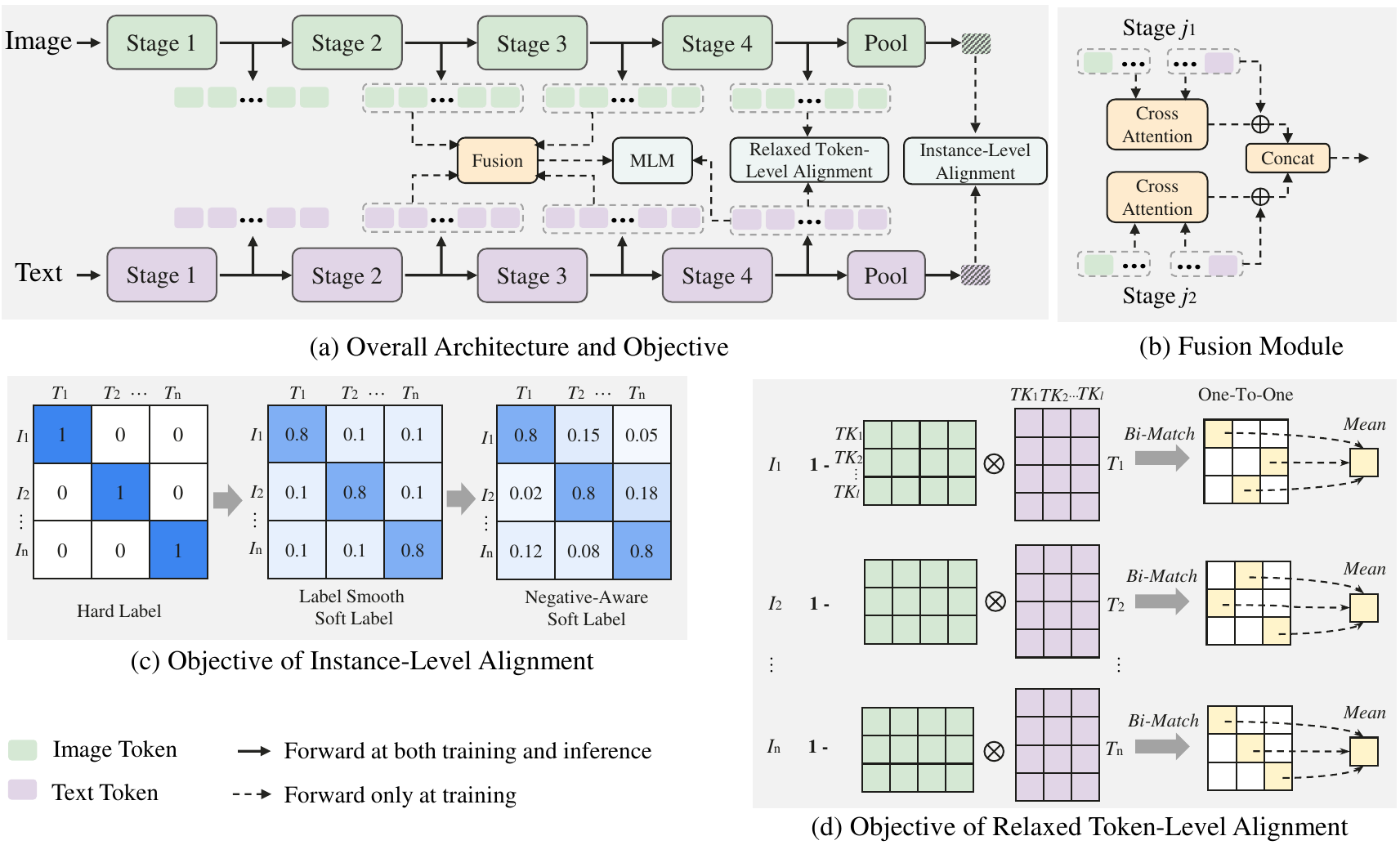}
	\caption{Illustration of LightCLIP. (a) Image and text encoder are each divided into four stages, the complete objective includes the objective of instance-level alignment, the objective of relaxed token-level alignment and MLM; (b) The fusion module of unmasked image to masked text for enhancing MLM; (c) Progressive label used in the objective of instance-level alignment; (d) The objective of relaxed token-level alignment, where $TK_i$ is $i$-th token, \textit{Bi-Match} is short for Bipartite-Matching. Only paired image-text is involved.}
	\label{fig_method}
\end{figure*}
In this section, we first introduce the overall architecture and objective of LightCLIP, then we elaborate the progressive softened instance-level alignment, the relaxed token-level alignment via bipartite matching and the multi-level fusion for enhancing MLM.

\subsection{Overall Architecture and Objective} 
The overall architecture and objective of LightCLIP are illustrated in Figure~\ref{fig_method}.(a), where the image and text encoder are each divided into four stages. Multiple lightweight networks including ResNet18~\cite{he2016deep}, MobileNet-V2~\cite{sandler2018mobilenetv2} and Swin-Nano~\cite{liu2021swin, dosovitskiy2020image} are adopted as our image encoders. The parameters of ResNet18, MobileNet-V2 and Swin-Nano are 12.51M, 2.45M and 7.11M, respectively. 8-layers Transformer~\cite{vaswani2017attention} with the architecture modifications described in~\cite{radford2019language} is adopted as our text encoder. For networks without stage divisions like Transformer in text encoder, we evenly divide the network into four stages according to its number of layers. 
The overall objective is formulated as:
\begin{equation}
\mathcal{L} =\alpha \mathcal{L}_{inst}+\beta \mathcal{L}_{token}+\gamma \mathcal{L}_{mlm}, 
\label{eq:loss}
\end{equation}
where $\alpha, \beta, \gamma \in [0,1]$ and $\alpha+\beta+\gamma=1$. $\mathcal{L}_{inst}$, $\mathcal{L}_{token}$ and $\mathcal{L}_{mlm}$ denote the objective of softened instance-level alignment, the objective of relaxed token-level alignment and the objective of enhanced MLM, respectively, and will be discussed in detail below. Following CLIP~\cite{radford2021learning}, $\mathcal{L}_{inst}$ is only computed at the output two-dimensional embedding of image and text. Empirically, $\mathcal{L}_{token}$ is calculated on the three-dimensional embedding at the fourth network stage. 

\subsection{Progressive Softened Instance-Level Alignment}
\noindent\textbf{Revisiting Instance-Level Alignment. }
For a image encoder $f$ and a text encoder $h$, given a batch of $n$ image-text pairs $\{(I_i, T_i)\}_{i=1}^n$, where $i$ indicates the $i$-th pair. The corresponding global embedding of image and text $\bm{v}_i \in \mathbb R^{d}$ and $\bm{t}_i \in \mathbb R^{d}$ are obtained separately by $\bm{v}_i = f(I_i)$ and $\bm{t}_i = h(T_i)$. In order to pull the embedding of paired image and text together while pushing unpaired apart, InfoNCE~\cite{oord2018representation} is widely adopted to perform the global alignment between the L2-normalized image and text embedding $\{(\bm{v}_i, \bm{t}_i)\}_{i=1}^n$. Specifically, for the $i$-th pair, the normalized image-to-text similarity $\bm{p}_{i}(I, T)=\{p_{ij}(I, T)\}_{j=1}^n$ and the text-to-image counterpart $\bm{p}_{i}(T, I)=\{p_{ij}(T, I)\}_{j=1}^n$ are calculated through:
\begin {gather}
p_{ij}(I, T) =\frac{\mathrm{exp}(\mathrm{sim}(\bm{v}_i,\bm{t}_j)/\tau)}{\sum_{j=1}^n \mathrm{exp}(\mathrm{sim}(\bm{v}_i,\bm{t}_j)/\tau)}, \\ 
p_{ij}(T, I) =\frac{\mathrm{exp}(\mathrm{sim}(\bm{t}_i,\bm{v}_j)/\tau)}{\sum_{j=1}^n \mathrm{exp}(\mathrm{sim}(\bm{t}_i,\bm{v}_j)/\tau)},
\label{eq:instanceloss1}
\end {gather}
where $\tau$ is a learnable temperature parameter initialized with $0.07$ and the function $\mathrm{sim}( \cdot )$ conducts dot product to measure the similarity score. In standard CLIP paradigm, the hard one-hot labels are used as the targets to calculate InfoNCE. Formally, let $\bm{y}_{inst}^{i}=\{y_{inst}^{i,j}\}_{j=1}^n$ denote the one-hot label of the $i$-th image-text pair, where $y_{inst}^{i,i}$ equal to $\mathrm{1}$ and all other values equal to $\mathrm{0}$. Then, the instance loss of image-to-text and text-to-image can be calculated by:

\begin {gather}
\mathcal{L}_{inst}^{i2t} =\frac{1}{n} \sum_{i=1}^n H(\bm{y}_{inst}^i, \bm{p}_{i}(I, T)),\\
\mathcal{L}_{inst}^{t2i} =\frac{1}{n} \sum_{i=1}^n H(\bm{y}_{inst}^i, \bm{p}_{i}(T, I)),
\label{eq:instanceloss2}
\end {gather}
where $H(\cdot, \cdot)$ represents the cross-entropy function. The final objective of global instance-level alignment is calculated through $\mathcal{L}_{\rm inst} = (\mathcal{L}_{inst}^{i2t}+\mathcal{L}_{inst}^{t2i})/2$.

\noindent\textbf{Progressive Softened Labels. }
However, as indicated in Figure~\ref{fig0}.(a), some image-text pairs are not strictly one-to-one correspondence. Therefore, the assumption that there is absolutely no similarity between unpaired image and text is defective. To mitigate this problem, PyramidCLIP~\cite{gao2022pyramidclip} proposes label smoothing by assigning weak weights to all the negative samples evenly:
\begin{align}
{\widetilde {\bm y} }_{inst}^i = (1-\delta) \times \bm{y}_{inst}^i + \frac{\delta}{n-1} \times (\mathbbm{1}-\bm{y}_{inst}^i),
\label{eq:instanceloss3_1}
\end{align}
where ${\widetilde {\bm y} }_{inst}^i$ represents the softened label vector, $\delta \in [0,1]$ is the hyper-parameter, and $\mathbbm{1}$ denotes the all-ones vector. However, label smoothing neglects the potential distinctions among negative samples. Therefore, we propose a negative sample importance-aware softening strategy:
\begin{equation}
{\hat {\bm y} }_{inst}^i = (1-\delta) \times \bm{y}_{inst}^i + \delta \times \mathrm{softmax}(\mathrm{fill}(\bm{y}_{inst}^i)),
\label{eq:instanceloss3}
\end{equation}
where ${\hat {\bm y} }_{inst}^i$ represents the softened label vector of ours, and 
\begin{equation}
\mathrm{fill}(\bm{y}_{i}) = 
\begin{cases}
-inf,& j=i\\
y_{inst}^{ij},& j\neq i
\end{cases}
\end{equation}
Eq.~\ref{eq:instanceloss3} can assign different weights to the negative samples according to their similarity scores, the higher the similarity score of the negative sample, the higher the assigned weight. Besides, to maintain training stability and achieve higher accuracy, we unify the above labels and introduce a progressive softening method. As illustrated in Figure~\ref{fig_method}.(c), as the pre-training progresses, the labels for calculating instance-level alignment objective gradually transition from the one-hot $\bm{y}_{inst}^{i} $ to the softened $\widetilde {\bm y} _{inst}^i$, and then to the softened ${\hat {\bm y} }_{inst}^i$. Formally, at the pre-training epoch of $e$, the labels for contrastive learning are formulated as:
\begin{equation}
\bm{y}_{inst}^{i} = 
\begin{cases}
\bm{y}_{inst}^{i},& e < r_1 \times E \\
\widetilde {\bm y} _{inst}^i, & r_1 \times E \leq e < r_2 \times E \\
{\hat {\bm y} }_{inst}^i, & e \geq r_2 \times E
\end{cases}
\label{eq:label}
\end{equation}
where $E$ represents the total epochs for pre-training, $r_1$ and $r_2$ denote the ratios for partitioning different pre-training epochs. Concretely, $r_1, r_2 \in [0,1]$ and $r_1 < r_2$.

\subsection{Token-Level Alignment via Bipartite Matching}
Instance-level alignment only calculate the similarity scores between image and text based on the global output two-dimensional embedding \ie $n\times d$, which neglects the finer local interactions between image patches and textual words (Figure~\ref{fig0}.(b)). To enhance the fine-grained representative ability of image and text encoder, the token-level information between image and text are required to align. There are some works implementing fine-grained interaction, but they usually introduce additional computational cost like off-the-shelf object detector~\cite{zhang2021vinvl,girshick2015fast} or text summary extractor~\cite{raffel2020exploring}.

To model the local interactions between image patches and textual words, while simultaneously maintain the efficiency, we propose an effective detector-free token-level alignment objective via bipartite matching. Given a cost matrix between two sets, such as the payoffs required for different workers to perform different jobs, bipartite matching aims to find a one-to-one correspondence between workers and jobs such that the overall payoff for completing all jobs is minimal. Similarly, when applying bipartite matching to the problem of token-level alignment between two modalities, two issues need to be resolved, \ie how to construct the two sets for correspondence and how to define the cost matrix . For the CLIP-alike two-stream paradigm, two separate embeddings from image and text naturally provide the two sets. Also, the similarity score computed by the function $\mathrm{sim}( \cdot )$ in Eq.~\ref{eq:instanceloss1} can be transferred to the calculation of cost matrix.

Considering the limited representative ability of the mobile image and text encoder, we align only the paired image-text and ignore the unpaired (Figure~\ref{fig_method}.(d)). Therefore, the learning difficulty of mobile image and text encoder is relatively reduced, forcing them to focus on learning the interaction between the paired image-text. Formally, different from the two-dimensional embedding utilized in instance-level alignment, token-level alignment is calculated based on three-dimensional embedding \ie $n \times l \times d$, where $n, l, d$ represent batch size, token number and channel, respectively. Denote $l_1$ and $l_2$ as the number of non-padded tokens of $i$-th image and $i$-th text, and the corresponding embeddings are $\bm{\mu}_i=f(I_i) \in \mathbb R^{l_{1}\times d}$ and $\bm{\omega}_i=h(T_i) \in \mathbb R^{l_{2}\times d}$. For the $i$-th image-text pair, the image-to-text cosine similarity $\bm{c}_{i}(I, T) = \{c_{i}^{s,t}(I,T)\}_{s=1,l_{1}}^{t=1,l_{2}}$ is calculated through:
\begin{equation}
c_{i}^{s,t}(I,T) = \frac{\bm{\mu}_i^{s} \cdot \bm{\omega}_i^{t}}{\left\|\bm{\mu}_i^{s}\right\|_{2} \left\|\bm{\omega}_i^{t}\right\|_{2}},
\label{eq:tokenloss1}
\end{equation}
It can be seen that $\bm{c}_{i}(I, T) \in [-1,1]$. Then, the $i$-th cost matrix of image-to-text is computed by:
\begin{equation}
\mathcal{L}_{match}^{i2t,i} = 1-\bm{c}_{i}(I, T), 
\end{equation}
where $\mathcal{L}_{match}^{i2t,i} \in \mathbb{R}^{l_{1}\times l_{2}}$ and $\mathcal{L}_{match}^{i2t,i} \in [0,2]$. Further, the optimal one-to-one matching $\hat{\sigma}_{i2t}^{i}$ between the $i$-th paired image-text with the lowest matching cost can be computed as:
\begin{equation} 
\hat{\sigma}_{i2t}^{i} = \argmin_{\sigma \in \mathfrak{S}_{l_{2}}} \sum_{s=1}^{l_{1}} \mathcal{L}_{match}^{i2t,i}[s, \sigma(s)],
\label{eq:tokenloss2}
\end{equation}
where $\mathfrak{S}_{l}$ denotes the set of all $l$ permutation. The lowest values in cost matrix $\mathcal{L}_{match}^{i2t,i}$ are obtained by the indexes of $[s, \sigma(s)]$. Hungarian~\cite{stewart2016end} algorithm is adopted to solve Eq.~\ref{eq:tokenloss2}. The token-level objective of image-to-text is then computed by:
\begin{equation}
\mathcal{L}_{token}^{i2t} =\frac{1}{n \times l_{1}} \sum_{i=1}^n \sum_{s=1}^{l_1} \mathcal{L}_{match}^{i2t,i}[s, \hat{\sigma}_{i2t}^{i,s}],
\label{eq:tokenloss3}
\end{equation}
Indeed, the token-level objective of text-to-image is equal to the token-level objective of image-to-text. Therefore, we directly get the final objective of token-level alignment by $\mathcal{L}_{\rm token} = \mathcal{L}_{token}^{i2t}$. 

Remarkably, our relaxed token-level alignment is only introduced at pre-training, therefore, the computational complexity is only $\mathcal{O}(nld)$ at pre-training and remains $\mathcal{O}(nd)$ at inference. Also, the bipartite matching scheme achieves a one-to-one correspondence between image patches and textual words with the lowest matching cost. On the contrary, the correspondence of one-to-many or many-to-one often occurs in other similar objector-free token-level alignment like FILIP~\cite{yao2021filip} and TokenFlow~\cite{zou2022tokenflow}, resulting in sub-optimal results and even negative effects for CNNs~\cite{cui2022democratizing}.

\subsection{Image to Text Fusion for Enhancing MLM}
As indicated in Figure~\ref{fig0}.(c), the accuracy of CLIP model does not increase correspondingly as the parameters of text encoder increase. Therefore, we compress text encoder by reducing its number of layers directly. Besides, to maximize the potential of the shortened text encoder, drawing on the successful experience of previous works~\cite{li2021align, li2021supervision, dou2022empirical}, the objective of masked language modeling (MLM) is leveraged. In detail, we first randomly choose 15\% of all tokens in each sequence. Following BERT~\cite{devlin2018bert}, the chosen tokens are then replaced by 10\% random tokens, 10\% unchanged, and 80\% \texttt{[MASK]}. Then, the embedding of the corresponding token is used to predict the original token. As illustrated in Figure~\ref{fig_method}.(a), our MLM is computed separately on two embedding: the embedding of the masked text itself and the fused embedding of unmasked image to masked text at multiple network stages. Intuitively, with the help of unmasked image embedding from image encoder, the learning of shortened text encoder will be easier and more efficient. Formally, consistent with Eq.~\ref{eq:tokenloss1}, we use $\bm{\hat{\omega}}_i=h(\hat{T_i}) \in \mathbb R^{l_{2}\times d}$ to denote the embedding of the fourth stage in text encoder with the $i$-th masked text $\hat{T_{i}}$ as input. The MLM calculated by only the embedding of text is first formulated as:
\begin{equation}
\mathcal{L}_{mlm}^{text} =\frac{1}{n} \sum_{i=1}^n H(\bm{y}_{mlm}^{i}, \mathrm{fc}(\bm{\hat{\omega}}_i)), 
\label{eq:mlmloss1}
\end{equation}
where $\bm{y}_{mlm}^{i}$ denotes the ground-truth one-hot vocabulary distribution of $i$-th text, $\mathrm{fc}$ represents the fully connected layer for predicting the original token. For the MLM with the embedding of unmasked image to masked text, empirically, fusion at multiple network stages tend to achieve higher performance. As illustrated in Figure~\ref{fig_method}.(b), take the fusion of two network stages as an example:
\begin{equation}
\begin{aligned}
&\underline{\bm{\hat{\omega}}}_i^{j_1} = \bm{\hat{\omega}}_i^{j_1} + \textsc{Cross-Att}(\mathrm{conv}(\bm{\mu}_i^{j_1}), \bm{\hat{\omega}}_i^{j_1}), \\
&\underline{\bm{\hat{\omega}}}_i^{j_2} = \bm{\hat{\omega}}_i^{j_2} + \textsc{Cross-Att}(\mathrm{conv}(\bm{\mu}_i^{j_2}), \bm{\hat{\omega}}_i^{j_2}), \\
&\underline{\bm{\hat{\omega}}}_i = \mathrm{concat}(\underline{\bm{\hat{\omega}}}_i^{j_1},\underline{\bm{\hat{\omega}}}_i^{j_2}), \\
&\mathcal{L}_{mlm}^{fuse} =\frac{1}{n} \sum_{i=1}^n H(\bm{y}_{mlm}^{i}, \mathrm{fc}(\underline{\bm{\hat{\omega}}}_i)), 
\end{aligned}
\label{eq:mlmloss2}
\end{equation}
where $\bm{\hat{\omega}}_i^j$ and $\bm{\mu}_i^j$ represent the masked text embedding and unmasked image embedding of $j$-th network stage, respectively. $\mathrm{conv}$ is short for the convolutional operation that transform the dimension of $\bm{\mu}_i$ to the dimension of $\bm{\hat{\omega}}_i$.  At last, the complete enhanced MLM is computed by $\mathcal{L}_{mlm} = (\mathcal{L}_{mlm}^{text}+\mathcal{L}_{mlm}^{fuse})/2$. Remarkably, the fusion module is only introduced at pre-training, therefore, the efficiency at inference can be maintained.

\section{Experiments} 

\subsection{Experimental Setup}
\noindent\textbf{Pre-training Datasets. }
In practice, the YFCC15M-V2 dataset~\cite{cui2022democratizing, li2021supervision} is adopted as the pre-training image-text pairs in our experiments, which is generated by filtering from the YFCC100M dataset~\cite{thomee2016yfcc100m} with a more meticulous strategy. Compared with other public datasets like CC3M~\cite{sharma2018conceptual},CC12M~\cite{changpinyo2021conceptual}, LAION-400M~\cite{schuhmann2022laion} and LAION-5B~\cite{schuhmann2022laion}, the medium-scale YFCC15M-V2 is a good balance of the training cost and performance. After removing the invalid download URLs, we end up with $15,255,632$ image-text pairs, slightly less than the original $15,388,848$ pairs in DeCLIP~\cite{cui2022democratizing, li2021supervision}.

\noindent\textbf{Implementation Details. }
Empirically, the hyper-parameters $\alpha$, $\beta$, $\gamma$ in Eq.~\ref{eq:loss} are set to 0.8, 0.1 and 0.1, respectively. $r_1$ and $r_2$ in Eq.~\ref{eq:label} are set to 0.33 and 0.66, respectively. $\delta$ in Eq.~\ref{eq:instanceloss3} and Eq.~\ref{eq:instanceloss3_1} is
\begin{table}[t]
	\captionof{table}{ZS classification results on ImageNet.}
	\makeatletter\def\@captype{table}
	\small
	\centering
	\setlength{\tabcolsep}{1mm}{\begin{tabular}{ccc}
			\toprule
			\textbf{Method} & \textbf{Image-Encoder} &  \textbf{Top1-Acc (\%)} \\
			\midrule
			CLIP & \multirow{4}{*}{ResNet18}& 30.8\\
			SLIP &  & 23.3   \\
			DeCLIP &  & 33.2   \\
			\textbf{LightCLIP} &  &  \textbf{35.0}   \\
			\midrule
			CLIP & \multirow{4}{*}{MobileNet-V2} & 28.0   \\ 
			SLIP & & 20.7  \\
			DeCLIP &  & 30.2 \\
			\textbf{LightCLIP} &  &  \textbf{31.8}   \\
			\midrule
			CLIP & \multirow{4}{*}{Swin-Nano} & 31.2  \\
			SLIP & & 31.3 \\
			DeCLIP &  & 33.7 \\
			\textbf{LightCLIP} &  &  \textbf{34.6}   \\
			\bottomrule
	\end{tabular}}
	\vspace{-1.0em}
	\label{tab1}
\end{table}
set to 0.2. $\mathcal{L}_{mlm}^{fuse}$ in Eq.~\ref{eq:mlmloss2} is calculated based on the embedding at the second and third network stages. To obtain better generalization and data-efficiency, data augmentation on both images and texts is performed. The images are augmented following the method in MOCO~\cite{chen2020improved} and SimCLR~\cite{chen2020simple}. The texts are augmented by random synonym replacement, random swap and random deletion. To save memory, automatic mixed-precision~\cite{micikevicius2017mixed} is used. Besides, the input images are resized to $224\times 224$ and the maximum length of the text is limited to 77 tokens following~\cite{radford2021learning}. We train our models using the AdamW~\cite{loshchilov2017decoupled} optimizer and the cosine learning rate scheduler with a linear warm-up. Specifically, the weight decay is set to 0.1, and the learning rate linearly increases from 0 to the peak value, \ie 5e-4 within 5\% of the total iterations. All our models are trained from scratch for either 8 or 32 epochs, \ie 8 epochs for ablation studies and 32 epochs for comparisons. In addition, 8 cards are used in our experiments.

\noindent\textbf{Downstream Tasks for Evaluation. } We evaluate on two downstream tasks: zero-shot image classification and zero-shot image-text retrieval. For zero-shot image classification, experiments are conducted on 11 datasets, such as ImageNet~\cite{deng2009imagenet}, CIFAR-10~\cite{krizhevsky2009learning}, Pets~\cite{parkhi2012cats}. For zero-shot image-text retrieval, experiments are carried out on Flickr30K~\cite{hodosh2013framing} and MS-COCO~\cite{lin2014microsoft}.

\begin{table*}[t]
	\setlength{\belowcaptionskip}{1pt}
	\centering
	\footnotesize
	\caption{Zero-shot accuracy on small datasets. AVG represents average accuracy across 10 datasets.}
	\label{tab2}
	\setlength{\tabcolsep}{0.8mm}{
		\begin{tabular}{cccccccccccccccc}
			\toprule
			\textbf{Image-Encoder} &\textbf{Method} &  \textbf{PETS} & \textbf{C10} & \textbf{C100} & \textbf{DTD} & \textbf{STL} &  \textbf{F101} & \textbf{FLOW} & \textbf{FER} & \textbf{SUN} & \textbf{CAL} &\textbf{AVG}\\ 
			\midrule
			\multirow{4}{*}{\tabincell{c}{ResNet18}} &CLIP   & 19.0 & 36.2 & 14.5 & 22.4 & 84.4 & 33.4 & 48.8 & 16.3 & 41.5 & 53.1  & 37.0    \\
			&SLIP   & 23.6 & 47.1 & 18.9 & 21.3 & 77.8 & 21.4 & 31.5 & 8.7 & 32.9 & 57.3 & 34.1 \\
			&DeCLIP   & \textbf{28.1} & 41.7 & 19.8 & \textbf{27.9} & 88.6 & 33.8 & 51.0 & \textbf{17.5} & 47.3 & 63.1 & 41.9 \\
			&\textbf{LightCLIP}  & 26.8 & \textbf{56.3} & \textbf{28.9} & 25.7 & \textbf{88.6} & \textbf{35.7} & \textbf{51.7} & 14.7 & \textbf{49.1} & \textbf{63.8} & \textbf{44.1} \\ 
			\midrule
			\multirow{4}{*}{\tabincell{c}{MobileNet-V2}} &CLIP   & 20.2 & 25.7 & 10.6 & 18.8 & 80.4 & 30.7 & 42.6 & 20.4 & 38.9 & 53.7  & 34.2    \\
			&SLIP   & 22.0 & 30.1 & 12.4 & 19.4 & 79.1 & 18.5 & 31.6 & 15.7 & 30.4 & 52.3 & 31.2 \\
			&DeCLIP   & 23.5 & 46.8 & \textbf{23.3} & 25.0 & 82.7 & 30.5 & \textbf{48.6} & \textbf{21.8} & \textbf{43.2} & \textbf{62.9} & 40.8 \\
			&\textbf{LightCLIP}  & \textbf{28.6} & \textbf{47.6} & 20.2 & \textbf{25.0} & \textbf{87.2} & \textbf{31.1} & 48.3 & 18.7 & 42.9 & 59.5 & \textbf{40.9} \\ 
			\midrule
			\multirow{4}{*}{\tabincell{c}{Swin-Nano}} &CLIP   & 22.4 & 47.2 & 20.6 & 21.2 & 87.1 & 31.3 & 47.6 & 19.3 & 41.0 & 55.5  & 39.3    \\
			&SLIP   & 24.2 & 58.3 & 26.7 & 21.6 & 86.1 & 34.4 & 50.0 & 17.3 & 41.6 & 59.6 & 42.0 \\
			&DeCLIP   & 25.3 & 61.2 & 32.8 & 24.7 & \textbf{90.4} & 32.6 & 49.7 & \textbf{21.0} & 47.9 & \textbf{66.4} & 45.2 \\
			&\textbf{LightCLIP}  & \textbf{29.2} & \textbf{71.0} & \textbf{37.4} & \textbf{26.0} & 89.6 & \textbf{37.4} & \textbf{52.1} & 16.6 & \textbf{48.1} & 64.5 & \textbf{47.2} \\ 
			\bottomrule
			\specialrule{0em}{1pt}{1pt}
	\end{tabular}}
\end{table*}

\begin{table*}[h]
	\footnotesize
	\caption{Zero-shot image-text retrieval results on Flicker30K and MS-COCO.}
	\label{tab3}
	\setlength{\belowcaptionskip}{2pt}
	\centering
	\setlength{\tabcolsep}{1.0mm}{
		\begin{tabular}{cccccccccccc}
			\toprule
			\multirow{5}{*}[5pt]{\tabincell{c}{\textbf{Image} \\ \textbf{Encoder}}} & \multirow{5}{*}[5pt]{\textbf{Method}} & \multicolumn{4}{c}{\textbf{Flickr30K}}  & \multicolumn{4}{c}{\textbf{MS-COCO}}  \\  
			\cmidrule(r){3-6}\cmidrule(r){7-10}
			& & \multicolumn{2}{c}{\textbf{Image-to-Text}}  & \multicolumn{2}{c}{\textbf{Text-to-Image}} & \multicolumn{2}{c}{\textbf{Image-to-Text}} & \multicolumn{2}{c}{\textbf{Text-to-Image}} & \\
			\cmidrule(r){3-4}\cmidrule(r){5-6}\cmidrule(r){7-8}\cmidrule(r){9-10}
			& & R@1 & R@5 & R@1 & R@5 & R@1 &  R@5 & R@1 & R@5  \\ 
			\midrule
			\multirow{4}{*}{ResNet18} &CLIP   & 37.4 & 69.3 & 27.4 & 54.1 & 24.5 & 50.3 & 16.2 & 36.9  \\ 
			&SLIP   & 29.2 & 59.5 & 22.3 & 47.9 & 18.6 & 41.9 & 12.9 & 31.4  \\
			&DeCLIP   & 47.5 & 75.8 & 31.0 & 58.9 & 26.3 & 50.9 & 16.2 & 37.1  \\
			&\textbf{LightCLIP} & \textbf{55.1} & \textbf{80.3} & \textbf{34.8} & \textbf{63.4} & \textbf{31.2} & \textbf{56.7} & \textbf{18.4} & \textbf{40.5} \\ 
			\midrule
			\multirow{4}{*}{MobileNet-V2} &CLIP   & 33.6 & 64.2 & 25.8 & 50.4 & 21.3 & 45.3 & 14.2 & 33.2  \\ 
			&SLIP   & 25.2 & 50.8 & 19.3 & 40.6 & 13.9 & 33.8 & 10.2 & 26.0 \\
			&DeCLIP   & 42.7 & \textbf{73.2} & 27.7 & 55.0 & 23.4 & 48.5 & 14.3 & 34.4  \\
			&\textbf{LightCLIP} & \textbf{45.1} &  73.0  & \textbf{29.4} & \textbf{57.3} & \textbf{27.6} & \textbf{51.9} & \textbf{15.9} & \textbf{36.0}  \\ 
			\midrule
			\multirow{4}{*}{Swin-Nano} &CLIP   & 40.7 & 72.2 & 27.8 & 55.0 & 24.7 & 49.4 & 16.0 & 36.5  \\ 
			&SLIP   & 41.5 & 74.5 & 29.2 & 56.6 & 25.8 & 51.3 & 15.4 & 35.7  \\
			&DeCLIP   & 49.1 & 75.3 & 30.8 & 58.8 & 25.6 & 50.8 & 15.9 & 36.5 \\
			&\textbf{LightCLIP} & \textbf{52.3} & \textbf{79.1} & \textbf{33.3} & \textbf{61.9} & \textbf{30.2} & \textbf{54.5} & \textbf{17.3} & \textbf{38.9}\\ 
			\bottomrule
			\specialrule{0em}{1pt}{1pt}
	\end{tabular}}
\end{table*}

\subsection{Zero-shot Image Classification}
\textbf{Zero-shot Accuracy on ImageNet.}
We first conduct experiments on the widely used zero-shot ImageNet classification task using the same amount of pre-training data YFCC15M-V2 in Table~\ref{tab1}. We compare LightCLIP with other methods including CLIP~\cite{radford2021learning}, SLIP~\cite{mu2022slip} and DeCLIP~\cite{li2021supervision}. The results of other methods are obtained by our implementation\footnote{based on the public code: https://github.com/Sense-GVT/DeCLIP}. Compared to CLIP baselines, our method improves the Top-1 accuracy by 4.2\%/3.8\%/3.4\% when the image encoder is ResNet18/MobileNet-V2/Swin-Nano, respectively. When compared to the competitive DeCLIP method, LightCLIP still maintains the advantage.

\textbf{Zero-shot Accuracy on Small datasets. }
We further report the zero-shot classification results on the other 10 small datasets, including Oxford-IIIT Pets~\cite{parkhi2012cats}, CIFAR-10~\cite{krizhevsky2009learning}, CIFAR-100~\cite{krizhevsky2009learning}, Describable Textures~\cite{cimpoi2014describing}, Stanford STL-10~\cite{coates2011analysis}, Food-101~\cite{bossard2014food}, Oxford Flowers-102~\cite{nilsback2008automated}, FER-2013~\cite{goodfellow2013challenges}, SUN397~\cite{xiao2010sun} and Caltech-101~\cite{fei2004learning}. The corresponding results are sequentially reported in Table~\ref{tab2}. Compared to CLIP baselines, our LightCLIP improves the average zero-shot accuracy by 7.1\%/6.7\%/7.9\% when the image encoder is ResNet18/MobileNet-V2/Swin-Nano, respectively.

\subsection{Zero-shot Image-Text Retrieval}
We evaluate LightCLIP on two retrieval benchmarks including Flickr30K~\cite{hodosh2013framing} and MS-COCO~\cite{lin2014microsoft}. The zero-shot results are presented in Table~\ref{tab3}. The experimental results demonstrate that LightCLIP brings continuous improvements with different image encoders. In particular, the top-1 hit accuracy improvements of image-to text on Flickr30K are significant, which can be attributed by our elaborate alignment objectives and network architecture during pre-training.
\begin{figure*}[t]
	\centering
	\includegraphics[width=0.95\linewidth]{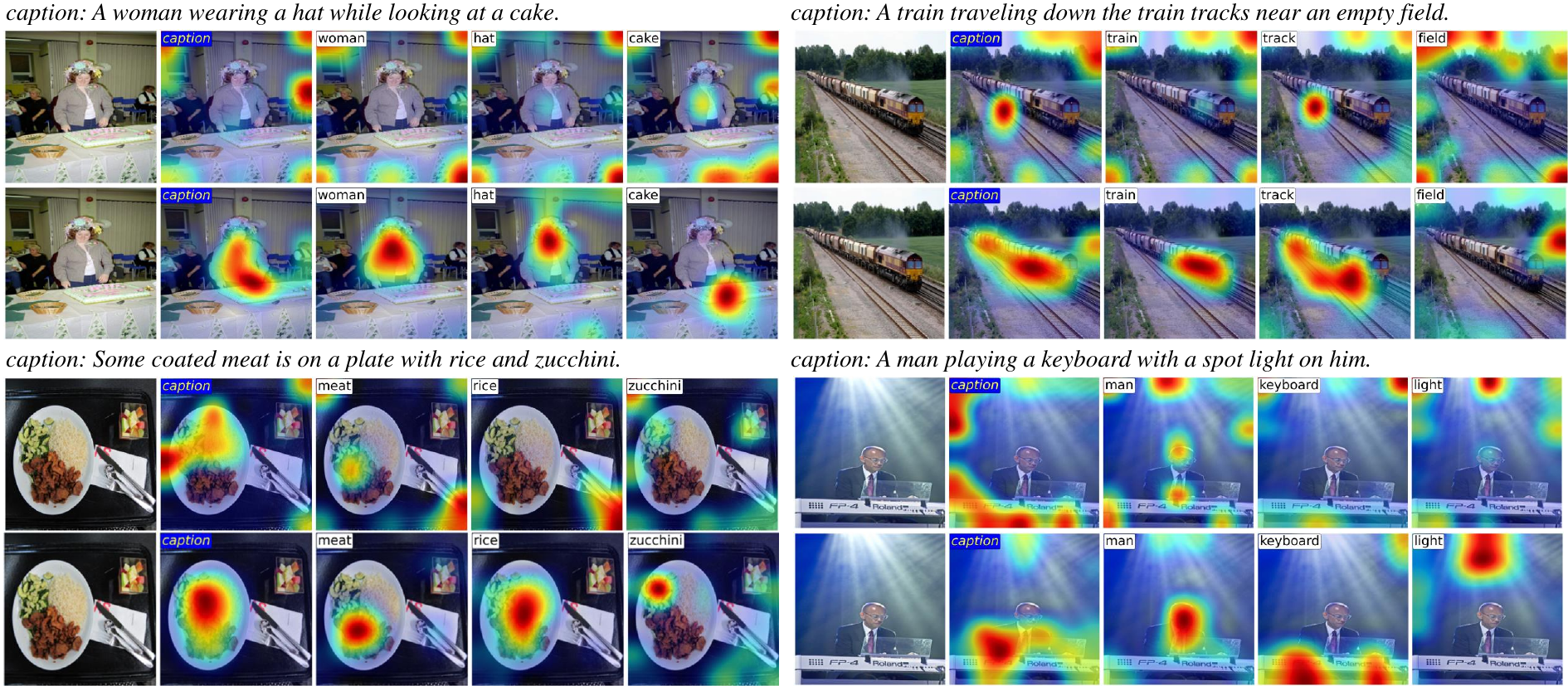}
	\caption{The correspondence between image patches and textual words on MS-COCO. In each example, the bottom is the proposed LightCLIP and the top is CLIP.}
	\label{fig_cam}
	\vspace{-1.0em}
\end{figure*}
\subsection{Ablation Study}
We conduct thorough ablation studies based on CLIP baseline. We analyze each component in Table~\ref{tab4}, where $Soft$ denotes the progressive softened labels. The introduction of each component brings an increase in accuracy, and the complete LightCLIP achieve the highest zero-shot accuracy for both networks.

\begin{table}[h]
	    \footnotesize
	    \caption{Analysis on each components.}
    	\label{tab4}
	    \setlength{\belowcaptionskip}{2pt}
    	\centering
		\setlength{\tabcolsep}{0.5mm}{\begin{tabular}{ccccc}
				\toprule
				\multirow{3}{*}[1.5pt]{\tabincell{c}{\textbf{Image}\\\textbf{Encoder}}} & \multicolumn{3}{c}{\textbf{Components}} & \multirow{3}{*}[1.5pt]{\tabincell{c}{\textbf{ImageNet}\\\textbf{ZS Top1}}} \\
				\cmidrule(r){2-4}
				& $Soft$ & $\mathcal{L}_{token}$ & $\mathcal{L}_{mlm}$  \\ \midrule
				\multirow{5}{*}{ResNet18} &  & &  &29.3 \\
				& \checkmark& & & 30.5 (\textcolor{darkergreen}{+1.2}) \\
				& & \checkmark& & 30.1 (\textcolor{darkergreen}{+0.8}) \\
				& & &\checkmark& 31.2 (\textcolor{darkergreen}{+1.9}) \\ 
				& \checkmark&\checkmark &\checkmark & 32.0 (\textcolor{darkergreen}{+2.7}) \\ 
				\midrule
				\multirow{5}{*}{Swin-Nano} &  & &  &29.5 \\
				& \checkmark& & & 30.4 (\textcolor{darkergreen}{+0.9}) \\
				& & \checkmark& & 30.2 (\textcolor{darkergreen}{+0.7}) \\
				& & &\checkmark& 30.8 (\textcolor{darkergreen}{+1.3}) \\ 
				& \checkmark&\checkmark &\checkmark & 31.7 (\textcolor{darkergreen}{+2.2}) \\ 
				\bottomrule
		\end{tabular}}
\end{table}

\subsection{Visualization}
\textbf{Grad-CAM Heatmaps. }
In Figure~\ref{fig_cam}, we visualize the correspondence between image patches and 
\begin{figure}
	\includegraphics[width=1.0\linewidth]{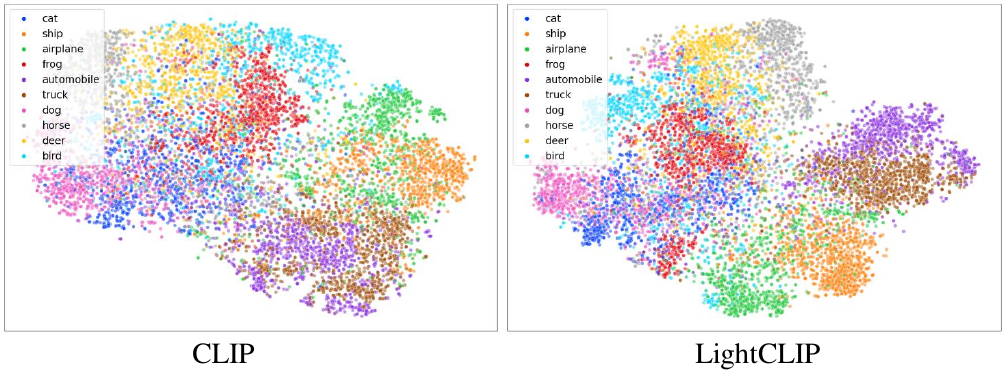}
	\caption{Image embeddings on CIFAR-10 test set.}
	\label{fig_tsne}
	\vspace{-1.0em}
\end{figure}
textual words with MobileNet-V2 image encoder by Grad-CAM heatmaps~\cite{selvaraju2017grad}. Obviously, LightCLIP achieves a more precise correspondence, which can explain the significant advantage of LightCLIP over CLIP on image-text retrieval task in Table~\ref{tab3}. For example, LightCLIP precisely capture the correspondence between the word \textit{cake} and the region in image, which is critical for downstream tasks.

\textbf{Image Embedding. }
We also visualize the extracted image embeddings with  Swin-Nano image encoder by \textit{t}-SNE~\cite{van2008visualizing}. As depicted in Figure~\ref{fig_tsne}, LightCLIP achieves a relatively better clustering performance, the embedding of most categories are more compact and highly differentiated, resulting in a higher classification accuracy.

\section{Conclusion}
In this paper, we explore mobile CLIP models under affordable computing resources. The many-to-many phenomenon in pre-training dataset, the failure correspondence between image patches and textual words, the problem that the parameters of text encoder increase without a corresponding increase in zero-shot accuracy \etc motivate the proposed LightCLIP. We first improve the traditional objective of global instance-level alignment by softening the label of negative samples progressively. Then, a relaxed bipartite matching based token-level alignment objective is devised for exploiting a finer-alignment. At last, the MLM enhanced by a multi-level fused embedding of unmasked image to masked text is leveraged for maximizing the potential of the shortened text encoder. Extensive experiments and ablation studies demonstrate the effectiveness of the proposed method.

\section{Acknowledgement}
We gratefully acknowledge the support of MindSpore~\cite{mindspore}, CANN(Compute Architecture for Neural Networks) and Ascend AI Processor used for this research. 

\bibliography{ref}
\bibliographystyle{icml2022}

\end{document}